\documentclass[10pt,twocolumn,letterpaper]{article}

\usepackage[pagenumbers]{cvpr}

\usepackage{graphicx}
\usepackage{amsmath}
\usepackage{amssymb}
\usepackage{booktabs}
\usepackage{pgfplots}
\usepackage{pgf-pie}
\usepackage{pgfplotstable}
\usepackage{multirow}

\usepackage{tabularx}
\usepackage{array}

\newcolumntype{Y}{>{\centering\arraybackslash}X}

\usepackage[pagebackref,breaklinks,colorlinks]{hyperref}

\usepackage[capitalize]{cleveref}
\Crefname{section}{Sec.}{Secs.}
\Crefname{section}{Section}{Sections}
\Crefname{table}{Tab.}{Tabs.}
\Crefname{table}{Table}{Tables}

\usepackage{soul}
\soulregister\cite7
\soulregister\cref7

\usepackage[accsupp]{axessibility}

\def\cvprPaperID{2}
\def\confName{CVPR}
\def\confYear{2022}

\pgfplotsset{compat=1.17}

\begin{document}

\title{A Challenging Benchmark of Anime Style Recognition}

\author{Haotang Li, Shengtao Guo, Kailin Lyu, Xiao Yang, Tianchen Chen\\
Jianqing Zhu\thanks{Corresponding authors}, and Huanqiang Zeng\footnotemark[1]\\\\
College of Engineering, Huaqiao University\\
Quanzhou, Fujian 362021, China\\
}
\maketitle

\begin{abstract}
Given two images of different anime roles, anime style recognition (ASR) aims to learn abstract painting style to determine whether the two images are from the same work, which is an interesting but challenging problem. Unlike biometric recognition, such as face recognition, iris recognition, and person re-identification, ASR suffers from a much larger semantic gap but receives less attention. In this paper, we propose a challenging ASR benchmark. Firstly, we collect a large-scale ASR dataset (LSASRD), which contains 20,937 images of 190 anime works and each work at least has ten different roles. In addition to the large-scale, LSASRD contains a list of challenging factors, such as complex illuminations, various poses, theatrical colors and exaggerated compositions. Secondly, we design a cross-role protocol to evaluate ASR performance, in which query and gallery images must come from different roles to validate an ASR model is to learn abstract painting style rather than learn discriminative features of roles. Finally, we apply two powerful person re-identification methods, namely, AGW and TransReID, to construct the baseline performance on LSASRD. Surprisingly, the recent transformer model (i.e., TransReID) only acquires a 42.24\% mAP on LSASRD. Therefore, we believe that the ASR task of a huge semantic gap deserves deep and long-term research. We will open our dataset and code at \url{https://github.com/nkjcqvcpi/ASR}.
\end{abstract}

\section{Introduction}
\label{sec:intro}

\begin{figure}[t]
    \centering
    \includegraphics[width=\columnwidth]{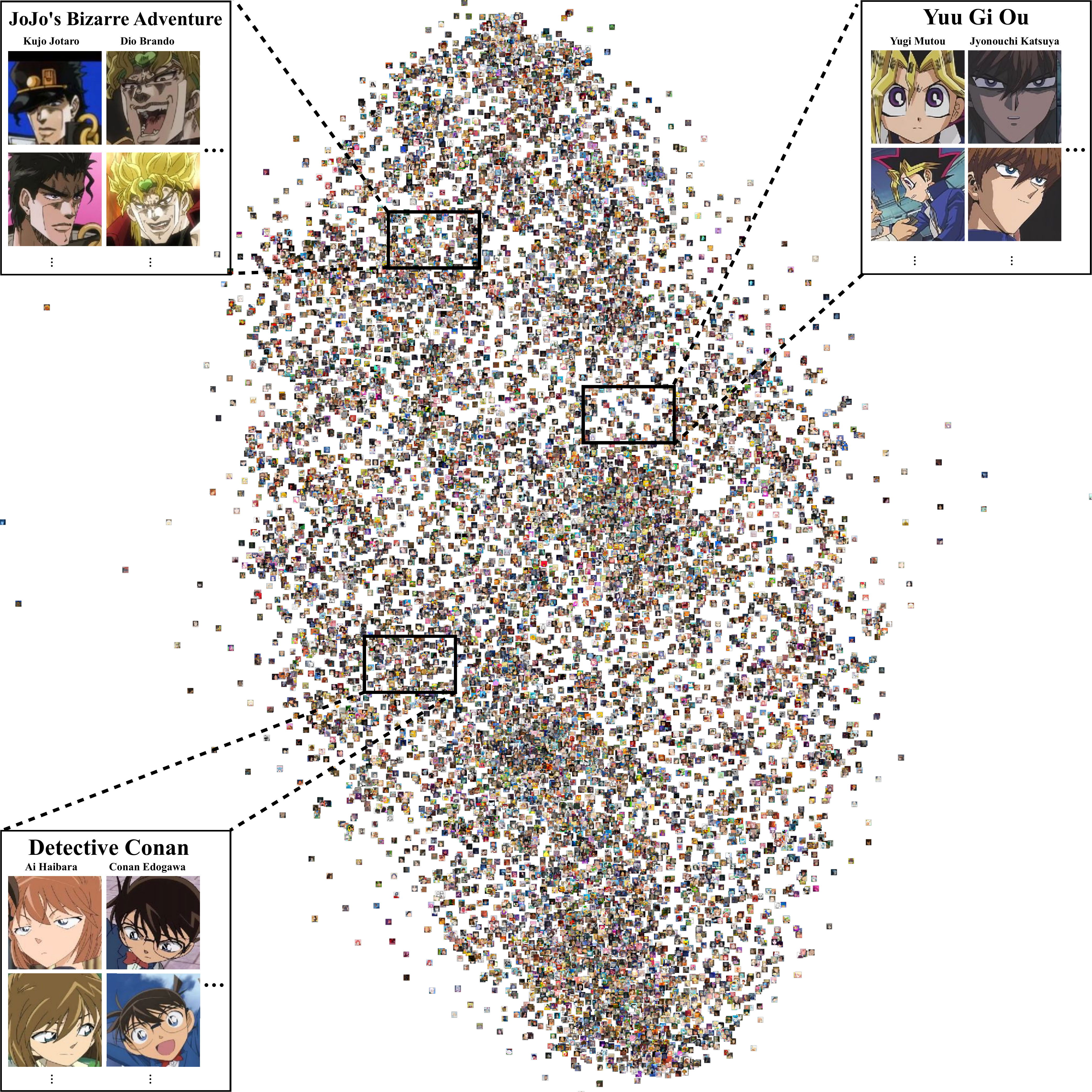}
    \caption{LSASRD Samples.}
    \label{fig:sample}
\end{figure}

Anime is a kind of animation products made in east Asia. A glance of anime images is shown in \Cref{fig:sample}. Anime was born in the last half-century and rapidly developed in the past 20 years. A large subculture group containing increasing audiences of anime worldwide is forming. Authors from various regions worldwide have different cultural backgrounds, which resulted in different painting styles in their anime works. With various styles and diverse themes, scenes of anime works could have more complex information than other forms of art or real-scene photos, which is an excellent carrier for research image understanding mechanisms of artificial intelligent. 

Human beings can recognize different faces by intuition \cite{face_brain}. With certain learning, they can conclude holistic style of different roles of one anime work \cite{prototype}. When given images of new roles, one experience can tell those roles are from the same works or not \cite{pattern_brain}. We are interest in whether computers can do the same thing. For that, we define an anime style recognition (ASR) task for computers, which learns the painting style of anime works to predict whether the images of various given roles are from the same work or not. 

\begin{table*}[t]
  \caption{Comparison of Datasets.}
  \centering
  \begin{tabularx}{\textwidth}{@{}lYYYYYY@{}}
    \toprule
    Dataset & iCartoonFace \cite{icartoonface} & Danbooru2021 \cite{danbooru2021} & Manga109 \cite{manga109} & COMICS \cite{comics} & nico-illst \cite{nico} & \textbf{LSASRD} \\
    \midrule
    Images & 389,678 & \~400k & 21,142 & \~1.2m & \~400k & \textbf{20,937}\\
    Roles  & 5,013 & Not annotated & \~500k & Not annotated & Not annotated & \textbf{1,829} \\
    Work   & Not annotated & Not annotated & 109 & Not annotated & Not annotated & \textbf{190} \\
    \midrule
    Applications & Face Detection / Recognition & Color Rendering & Text Detection & Plot & Color Rendering & \textbf{Style Recognition} \\
    \bottomrule
  \end{tabularx}
  \label{tab:compares}
\end{table*}

Distinctive from usual biometric recognition \cite{biometric_recognition,agw,face_recognition,vit,iris_recognition} tasks paying attention to learning identity context, ASR focuses on learning painting styles of anime works. Specifically, most common biometric recognition, e.g., face \cite{face_recognition} recognition and iris \cite{iris_recognition} recognition, one category contains multiple images. But, in ASR, one category is work includes multiple roles and one role consists of multiple images. Thus, ASR is a more challenging computer vision task, which suffers from complex intra-class variations.
In addition to biometric recognition, some researchers have made progress in understanding the arts. For example, Mao \etal \cite{deepart} proposed a DeepArt method to learn joint representations that can simultaneously capture contents and style of visual arts. Garcia \etal \cite{Text2Art} proposed a multi-modal retrieval method, namely, Text2Art, to retrieve arts by text queries. Shen \etal \cite{spatially-consistent} designed a spatially-consistent feature learning method to discover visual patterns in art collections. Some research on images style focuses more on descriptions \cite{styletexture} or classification \cite{styleclassific} on a single sample but a bunch of images with rich contextual information. However, anime could not be simply treated as a kind of fine art. Because anime is free in designing that contains complex illuminations, various poses, theatrical colors and exaggerated compositions.

\begin{figure}[t]
    \centering
    \includegraphics[width=\columnwidth]{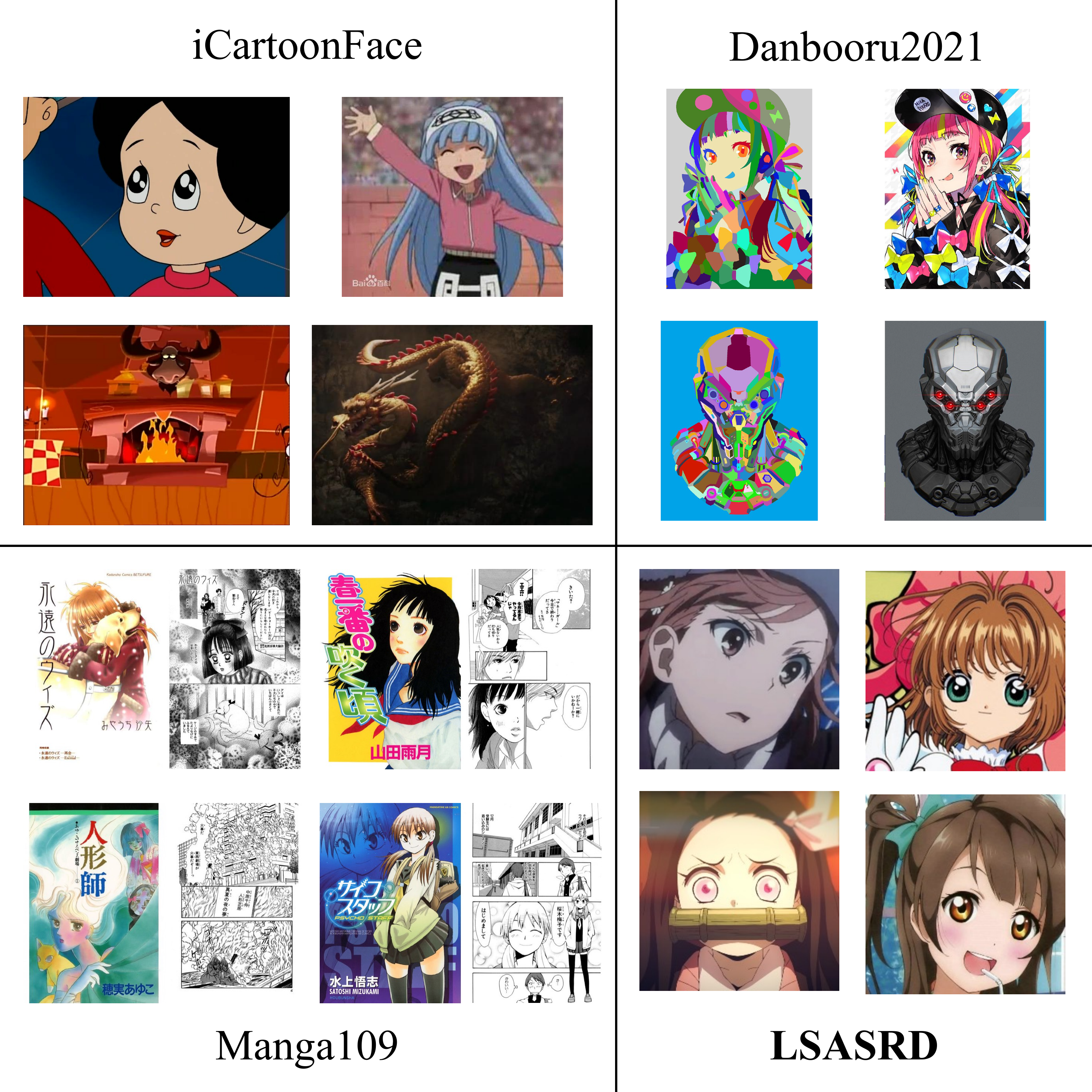}
    \caption{Examples of different datasets.}
    \label{fig:compares}
\end{figure}

Some anime image datasets \cite{icartoonface, danbooru2021, nico} have been released recently. For example, Aizawa \etal \cite{manga109} proposed the Manga109 dataset, which only contains scanned images of comic books and each image's text boxes and comic panels are labeled, so that Manga109 is especially for text detection and image splitting. Zheng \etal \cite{icartoonface} proposed the iCartoonFace dataset for cartoon face detection and recognition. Although iCartoonFace has a great number of images, many images are consecutive frames a cartoon video, causing many images have similar content even miss faces. Ikuta \etal \cite{nico} proposed the Nico-illust dataset, which mainly is composed of painting images from \href{http://seiga.nicovideo.jp/}{Niconico Seiga}. Nico-illust is designed for anime color rendering. \Cref{fig:compares} and \Cref{tab:compares} give a brief comparison. As as result, a ASR dataset richly annotated role and work information is still needed at present.

In this paper, an anime-style recognition dataset (LSASRD) is proposed. LSASRD contains 20937 face images of 1829 roles in 190 anime works between 1928 and 2021. Then, a cross-role protocol is designed by strictly dividing query set and gallery set with no same label of roles existing in both sets simultaneously, which enforces a deep learning model to holistically learn painting styles rather than identity contexts. Moreover, we apply mean inverse negative penalty (mINP) \cite{agw}, mean average precision (mAP) \cite{map} and cumulative matching characteristics (CMC) \cite{cmc} curves to evaluate the ASR performance. Finally, we use two state-of-the-art person re-identification methods, namely, AGW \cite{agw} and TransReID \cite{vit} to construct baseline ASR performance. Surprisingly, we found that existing methods performed poorly on LSASRD, in which AGW performs 40.84\% of mAP, and TransReID performs 42.76\% of mAP, which shows that ASR is challenging and requires much research to be done. And with a few focus on the style of anime images at present, our research on anime-style images will benefit the development of information recommendations and image retrieval about the anime content. 

This paper is organized as follows. \Cref{sec:dataset} describes the detail of LSASRD. \Cref{sec:protocol} designs a cross-role protocol to evaluate ASR. \Cref{sec:baseline} introduces two baseline ASR methods. \Cref{sec:experiment} reports experiment results and analysis. \Cref{sec:con} concludes this paper.

\section{Related Work}
\label{sec:related}

\subsection{Anime Recognition}
Recent research on anime is more likely to develop image generation \cite{animegan}, colorization \cite{cartoonize}, style transfer \cite{animestyletransfer} or video interpolation\cite{interpolation} methods. Recognition \cite{mangarecognition} and detection \cite{animefacedetector} on anime roles has gradually received this year. Several datasets have been put out, \eg Zhang \etal~\cite{icartoonface} present a new benchmark dataset named iCartoonFace, And provide two types of annotations, face recognition and face detection. Zhang \etal~\cite{danbooregion} make an Illustration Region dataset, this dataset named DanbooRegion., which contains a lot of artistic region compositions paired with corresponding cartoon illustrations. Aizawa \etal~\cite{manga109} built a dataset, which is named Manga109. This dataset provides numerous manga images and annotations. Zhang \etal~\cite{mangagan} make a framework to generate manga from digital illustration and present a data-driven framework to Convert the illustration into three components. Iyyer \etal~\cite{comics} design a dataset, which is named COMICS. Moreover, present three close-style tasks. For other research of anime, LSASRD focuses on the problem of identification. However, unlike conventional recognition problems, LSASRD does not just categorize the same role. It also needs to categorize different roles from the same anime.

\subsection{Face Recognition}
Face recognition is widely regarded as a sub-problem of image retrieval. Face recognition aims at extracting different features of a human face for recognition. With the development of face recognition, it is a very effective method to extract facial features for classification. Therefore, we choose to extract the facial features of roles in the experiment. The features of face recognition are also summarized. \eg, Liu \etal~\cite{celebfaces} introduce the CelebFaces Attributes (CelebA) dataset. This dataset has more than 200K images, each with 40 attribute annotations. Nowadays, face recognition is mainly collected in real life. For these datasets, the proposed methods can achieve very advanced results on the classification of these datasets. In anime, usual feature extraction methods and points may differ from real datasets. Human to abstract humanoid is a tremendous challenge for models.

\subsection{Person Re-identification}
Person Re-identification aims to use computer vision technology to determine a specific pedestrian in an image or video sequence, which needs to solve the difficulties \eg complex illumination variations, viewpoint or pose. With many methods of Person Re-identification have been proposed which has significantly been developed as it is being more and more attention, some difficulties have been effectively resolved. Both CNN and Transformer neural networks are used for Person Re-identification. \eg Ye \etal~\cite{agw} make a strong AGW baseline. It can reach state-of-the-art or at least comparable performance on both single- and cross-modality Re-ID tasks. Besides, \eg Ye \etal~\cite{agw} propose an evaluation criterion mINP for Person Re-identification. He \etal~\cite{transreid} introduced a new transformer-based object ReID framework, And designed two modules, the jigsaw patch module and the side information embeddings. These two methods have achieved SOTA results. 


\section{Large-Scale Anime Style Recognition Dataset}
\label{sec:dataset}

\pgfplotsset{
    compat=1.9,
    compat/bar nodes=1.8,
}
\pgfplotstableread{
    Label China Japan U.S. U.K. France Korea Other
    1920- 0 0 2 0 0 0 0
    1950- 5 0 0 0 0 0 1
    1960- 10 2 0 0 0 0 0
    1970- 5 3 6 2 0 0 2
    1990- 5 5 9 1 1 0 1
    2000- 13 16 10 2 4 0 1
    2010- 24 36 16 1 1 4 2
    2020- 5 4 2 0 0 0 0 
}\testdata

\begin{figure*}[t]
    \centering
    \begin{subfigure}{\columnwidth}
        \begin{tikzpicture}
            \begin{axis}[
                ybar stacked,
                ymin=0,
                ymax=90,
                width=1.05\linewidth, 
                height=150, 
                xtick=data,
                legend style={
                    cells={anchor=west},
                    legend pos=north west,
                },
                reverse legend=true,
                xticklabels from table={\testdata}{Label},
                xticklabel style={align=center}, 
            ]
            \addplot [fill=green!60] table [y=China, meta=Label, x expr=\coordindex]
                {\testdata}; \addlegendentry{China}
            \addplot [fill=blue!60] table [y=Japan, meta=Label, x expr=\coordindex]
                {\testdata}; \addlegendentry{Japan}
            \addplot [fill=red!60] table [y=U.S., meta=Label, x expr=\coordindex]
                {\testdata}; \addlegendentry{U.S.}
            \addplot [fill=yellow!60] table [y=U.K., meta=Label, x expr=\coordindex]
                {\testdata}; \addlegendentry{U.K.}
            \addplot [fill=pink!60] table [y=France, meta=Label, x expr=\coordindex]
                {\testdata}; \addlegendentry{France}
            \addplot [fill=gray!60] table [y=Korea, meta=Label, x expr=\coordindex]
                {\testdata}; \addlegendentry{Korea}
            \addplot [fill=black!60] table [y=Other, meta=Label, x expr=\coordindex]
                {\testdata}; \addlegendentry{Other}
            \end{axis}
        \end{tikzpicture}
        \caption{Year \& Region}
        \label{fig:statistics-a}
    \end{subfigure}
    \begin{subfigure}{0.5\columnwidth}
        \begin{tikzpicture}[scale=0.69]
            \pie[text=inside, color={yellow, green, cyan}]{67.9/Male, 22.8/Female, 9.3/Other}
        \end{tikzpicture}
        \caption{Gender}
        \label{fig:statistics-b}
    \end{subfigure}
    \begin{subfigure}{0.5\columnwidth}
        \begin{tikzpicture}[scale=0.69]
            \pie[text=inside]{69.2/Human, 17.0/Humanoid, 13.8/Inhuman}
        \end{tikzpicture}
        \caption{Race}
        \label{fig:statistics-c}
    \end{subfigure}
    \caption{Distribution of time and regions in LSASRD.}
    \label{fig:statistics}
\end{figure*}
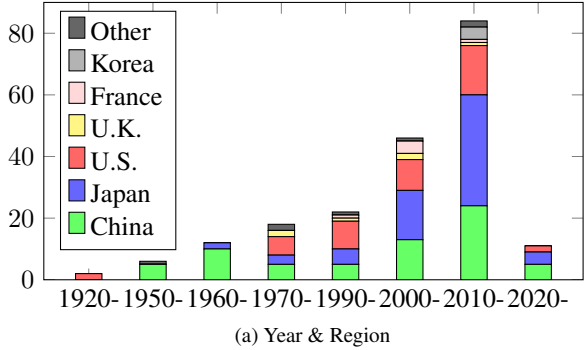
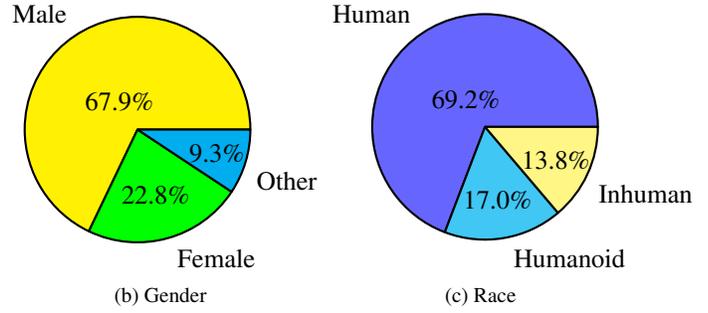

\subsection{Overview}
We present a well-labeled challenging dataset, Large Scale Anime Style Recognition Dataset (LSASRD), to facilitate the research on style recognition on anime images by collecting images from 190 anime and cartoon works covering 93 years from 13 countries and regions, 2D and 3D work into consideration concurrently. We choose at most ten roles for each work. All the images are obtained from the Internet. 
The images in the LSASRD dataset are mainly from existing anime and cartoons. Moreover, some are from comics or games of the same anime series. Unlike illustration or video datasets, we provide a moderate amount of contextual information in a wide variety of styles. LSASRD requires the ability of context understanding of image models.
To embody our target of recognition of styles of anime works, we selected works from 13 countries and regions between 1928 and 2021, briefly shown in \Cref{fig:statistics-a}. Two-thirds of works are from China and Japan, and 39\% were broadcast between 2010 and 2020. To differentiate from humanoid roles we are familiar with, we select several inhuman roles especially. In the statistics, 58.3\% of roles in LSASRD are male, and 29.5\% are female, as \Cref{fig:statistics-b}. 69.2\% of roles are human, 16.9\% is humanoid such as furry, and 13.8\% is inhuman, such as animal, as \Cref{fig:statistics-c}.


\begin{figure}[htb]
    \centering
    \includegraphics[width=\columnwidth]{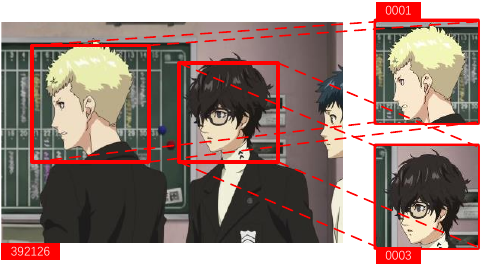}
    \caption{Annotations of LSASRD.}
    \label{fig:labeling}
\end{figure}

\subsection{Data Collection and Annotation}
We fetch a list of anime works from \href{https://zh.moegirl.org.cn}{Moegirlpedia} and \href{https://www.bilibili.com}{BiliBili}. Sorting by popularity, regarding the distribution of eras and regions, we choose 190 works by weighted randomly. We also consider it the same work for some serial spans an extended period or have many branches works. Then for the roles, we chose both the main characters and supporting roles of the work. While some work contains even less than ten roles, we have to reduce the number of labels. Finally, 1829 labels are presented in our dataset. While the images on the Internet of main roles significantly more than supporting ones, we thus construct a semantic environment that under different dataset partition conditions, the model is requested to learn the style from both lot-shot labels and few-shot labels, which simulate the tasks in the actual scene. We get images from picture searching engines and various online video websites in the public domain or under fair use. Unprocessed data contain 60k images of uneven quality, covering manga, comics, episodes, movies, even photos of peripheral products. Secondary creations are also under our consideration. We mark the bounding boxes and label the images manually using a self-development tool. Five collaborators clean and annotate the data for about two months; 20937 images are selected and well-labeled.

In order to control the uncertainties of the scenes and technical factors of drawing, we only use the images of the face part of each role. Put another way, images in LSASRD currently all are the portrait of roles. Moreover, the whole origin image can be used in more fields, i.e., detection and recognition. We plan to use them in further research of high-level semantics of entities beyond the roles. All images are cropped and only show head of each role, resized in 256px by 256px, as \Cref{fig:labeling} shows. We label the picture by the title of work and mark the face in the images using bounding boxes and labels of whom the faces belong. 
We also construct the metadata of each work, role, and image handcrafted, \eg year, region, staff of works and the identity information \eg gender, race of roles, to provide context information as rich as possible.

\begin{figure*}[t]
    \centering
    \includegraphics[width=\linewidth]{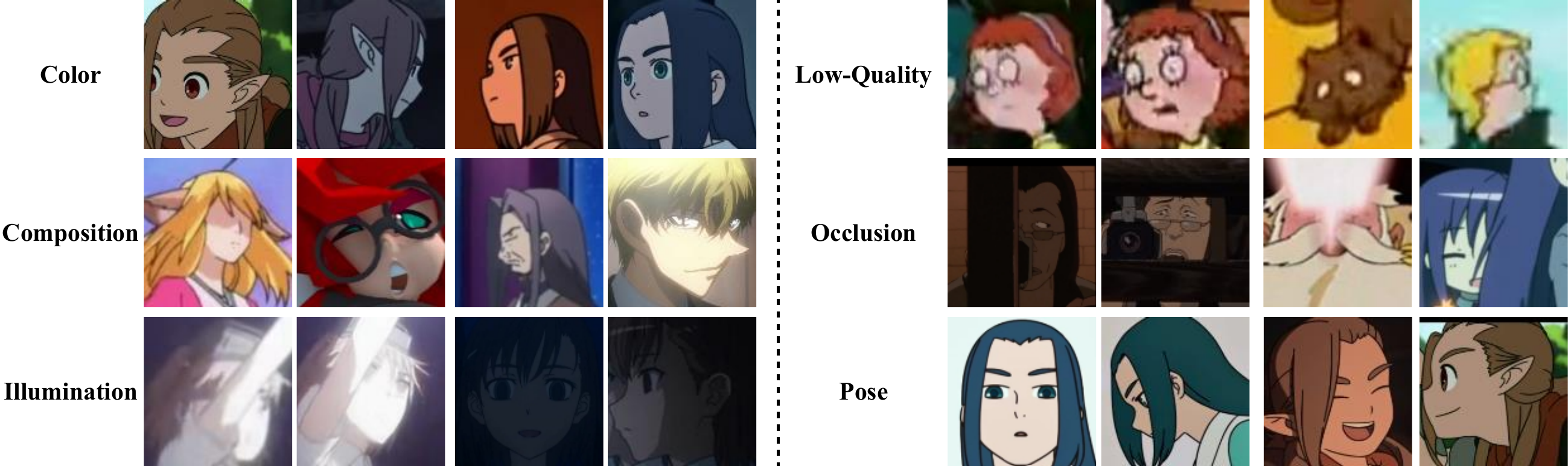}
    \caption{Examples of LSASRD.}
    \label{fig:challenges}
\end{figure*}

\subsection{Challenge}

\textbf{Complex image content conditions.} A part of pictures in lower clarity, too bright or too dark, large-area occlusion, exaggerated pose, and special composition distribute uniformly in the dataset. Various resolutions, colors, illumination, angles, or keep outs of objective factors build the typical difficulties of image data. The samples of typical challenges are shown in \Cref{fig:challenges}.

\textbf{Complex image style conditions.} Focus on huge semantics gap recognition, images in LSASRD of a role selected in both the original work and secondary creation works, which means that the level of roles could contain different styles. The images are from multiple regions and eras with different styles and roles of human beings, humanoid and inhuman, which means that model can hardly learn the inherent patterns of facial features of a kind of role. Moreover, similar roles from different works and dissimilar roles from the same works are considered the difficulties of the semantics challenge, which is the core of our research.

\textbf{More complicated than biometric recognition datasets.} One subject in biometric recognition datasets \cite{celebfaces, iris_dataset} commonly contains different images of the same person, such as different shots captured by different cameras for one person, different expressions on the same face. However, LSASRD is much more complex, in where one subject means an anime work of different roles and each role also of different images in different status. For models, they cannot just learn the identity of images because the identities are different in roles. Unlike the conception of cameras or views, roles can be considered different aspects of a machine to understand semantics between each works. They have to look for high-level common features.


\section{Cross-Role Protocol}
\label{sec:protocol}
To evaluate the performance of model of understanding the style differentiation in LSASRD. The details of our protocol are described as follows.

\subsection{Training Set and Testing Set Dividing}
For our publishing LSASRD, images are randomly divided into two-part for training and testing, as shown in \Cref{tab:divide}. Further, the test set is divided into query set and gallery set by the roles of each work and the ratio of 6:4. Notably, the roles of pictures in the query set are strictly do not exist in the gallery set. This cross-role data dividing allows for our protocol to be a cross-role protocol. 

\begin{table}[htb]
    \caption{Splitting of LSASRD.}
    \centering
    \begin{tabular}{@{}lcccc}
        \toprule
        Subset & Train & Gallery & Query \\
        \midrule
        Work & 114 & 76 & 76\\
        Role & 1097 & 439 & 293\\
        Image & 12562 & 5025 & 3350 \\
        \bottomrule
    \end{tabular}
    \label{tab:divide}
\end{table}

In order to reduce the bias from data distribution and overfitting, we introduce 5-fold cross-validation to build our baseline dataset. It is a type of k-fold cross-validation when k = 5. A single 5-fold cross-validation is used with both a training and validation set. The total data set is split into five sets as \Cref{tab:devide_cv} shows. One by one, a set is selected as the validation set. Then, the other four sets are used as training sets until all possible combinations have been evaluated. The training set is used for model fitting. The validation set is used for model evaluation for each hyper-parameter set. The mean value of each five evaluations calculates our baseline performance.

\begin{table}[htb]
    \caption{5-Fold Cross-Validation Setting.}
    \centering
    \begin{tabular}{@{}lcccccc}
        \toprule
        Fold-$ k $ & 1 & 2 & 3 & 4 & 5\\
        \midrule
        Work & 38 & 38 & 38 & 38 & 38\\
        Role & 364 & 367 & 367 & 366 & 365 \\
        Image & 4187 & 4185 & 4188 & 4189 & 4188 \\
        \bottomrule
    \end{tabular}
    \label{tab:devide_cv}
\end{table}

\begin{figure*}[t]
    \centering
    \includegraphics[width=0.9\linewidth]{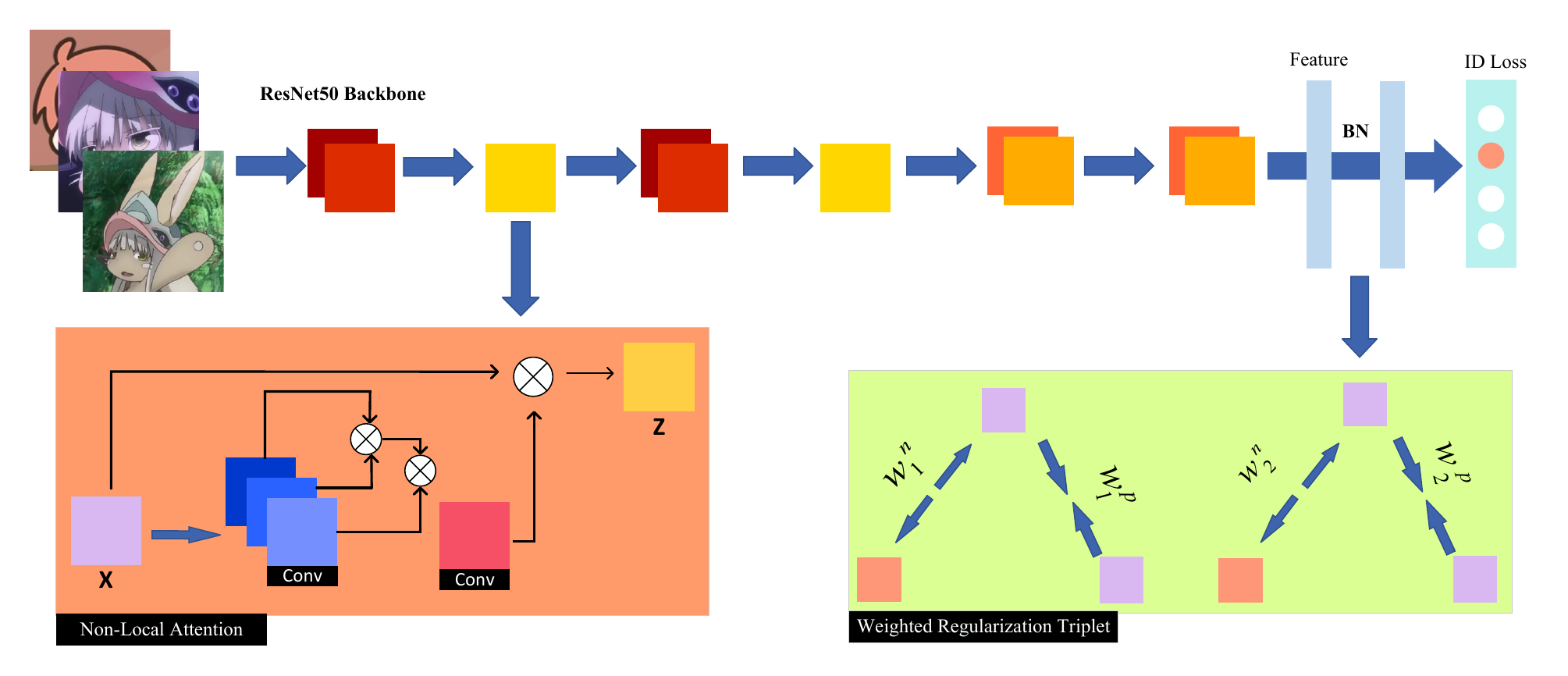}
    \caption{AGW framework \cite{agw}}
    \label{fig:AGW}
\end{figure*}

\subsection{Performance Metric}

In order to evaluate the ability of model to learn semantics, we have considered several methods, such as simple classification or clustering. Finally, we choose Re-identification based on deep metric learning. In the Re-identification process, a candidate list sorted by the feature distances between the query and gallery images is returned from the dataset for each given query. The mean Inverse Negative Penalty (mINP)  \cite{agw}, mean Average Precision (mAP) \cite{map} and Cumulative Matching Characteristics (CMC) \cite{cmc} are used as performance metrics.

The mINP evaluates the ability to retrieve the hardest correct match, providing a supplement for measuring the Re-ID performance, which is formulated as \cref{eq:mINP}.
\begin{equation}
    mINP = \frac{1}{n} \sum_{i}(1-NP_{i}) = \frac{1}{n} \sum_{i} \frac{|G_{i}|}{R_{i}^{hard}},
    \label{eq:mINP}
\end{equation}
where $ NP $ measures the penalty to find the hardest correct match, and calculated as $ NP_i = \frac{R_i^{hard} - |G_i|}{R_i^{hard}} $, where $ R_i^{hard} $ indicates the rank position of the hardest match, and $ |G_i| $ represents the total number of correct matches for query $ i $.

The mAP calculate the mean value of average precision scores for each batch, which is formulated as \cref{eq:mAP}.
\begin{equation}
    mAP = \frac{\sum_{q = 1}^Q AP(q)}{N_{Q}}
    \label{eq:mAP}
\end{equation}
where $ N_Q $ is the number of batch, and $ AP $ is calculated as $ AP = \frac{\sum_{k = 1}^{n}P(k)\times rel(k)}{N_{Relevant}} $, where $ k $ is the rank of a size $ n $ recall list, $ N $ is the number of relevant works. $ P(k) $ is the precision at cutoff, and $ rel(k) $ indicates whether the $ k $-th recall is correct or not. $ N_{Relevant} $ is the number of total query images. In addition, the top-$ k $ match rate is also reported in experiments. The average value of the precision function of recall is the Average Precision instead of the single-value metrics precision and recall based on each query and gallery. mAP presents an overall performance for re-identification.

The CMC curves show the probability of a query identity appearing in different-sized candidate lists. The cumulative match characteristics at rank $ k $ can be calculated as \cref{eq:CMC}.
\begin{equation}
    CMC@Rank(k) = \frac{\sum_{q=1}^{Q}rel(q, k)(k)}{Q},
    \label{eq:CMC}
\end{equation}
where $ rel(q, k) $ is an indicator function equaling one if the ground-truth of $ q $ image appears before rank $ k $ of gallery images, zero otherwise.


\begin{figure*}[htb]
    \centering
    \includegraphics[width=\linewidth]{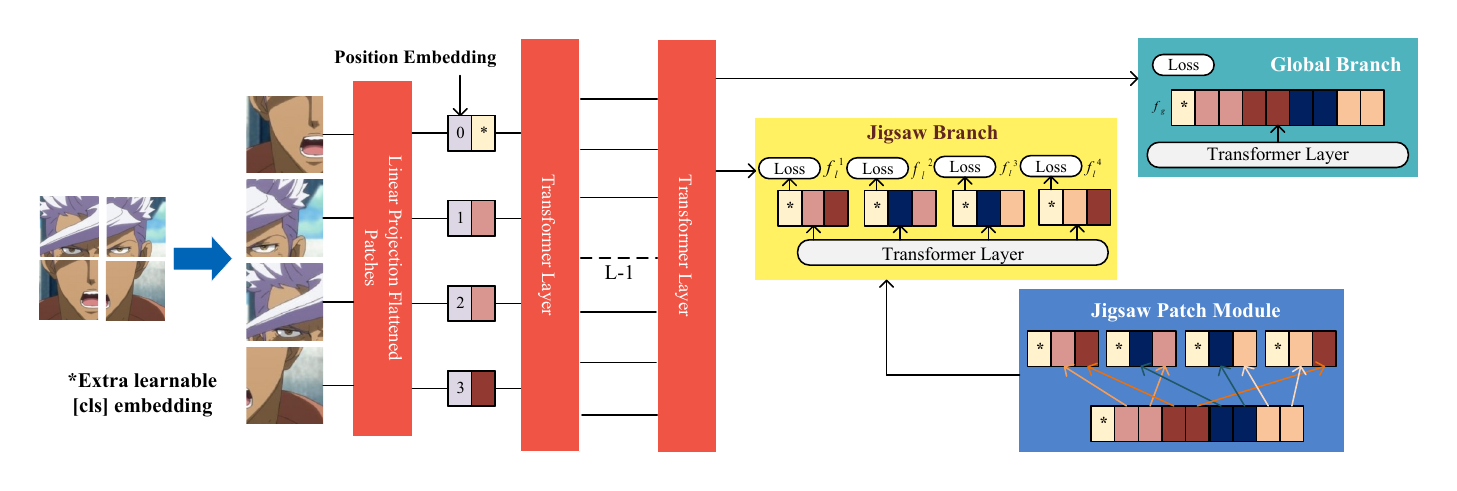}
    \caption{TransReID framework \cite{transreid}.}
    \label{fig:TransReID}
\end{figure*}

\section{Baseline Methods}
\label{sec:baseline}

As shown in \Cref{fig:AGW}, AGW includes backbone network structure, criterion and the hyper-parameter configuration for ResNet50 with non-local block (ResNet50 NL).
For backbone, ResNet50 NL is the default, we also use pure ResNet with 50, 101 and 152 layers. Instance-batch normalization (IBN) leads to build the ResNet50 IBN A model. For a better comparison, we introduce squeeze and excitation (SE) modules in ResNet to construct SE ResNet with 50, 101 and 152 layers. Further, ResNext with SE module pulls in to build the SE ResNext50 and 101.

For criterion, label smoothing cross-entropy and weighted regularized triplet are used to calculate the loss. Label smoothing \cite{labelsmooth} encourages the differences between the logits of the correct class and the logits of the incorrect classes to be a constant. Weighted regularization inherits the advantage of relative distance optimization between negative and positive pairs but avoids introducing any additional margin parameters.

For the TransReID shown in \Cref{fig:TransReID}, the base and small size of Vision Transformer (ViT) \cite{vit} and data-efficient image transformers (DeiT) \cite{deit} are used as backbone, namely ViT-Base, ViT-Small and DeiT-Small. Jigsaw Patch Module (JPM) proposes rearranging the patch embeddings via shift and patch shuffle operations, generating robust features with improved discrimination ability and more diversified coverage. With JPM, we set an experiment of ViT-JPM. With default Transformer stride of 16, we set a group of stride of 12, namely ViT-Stride and DeiT-Stride.
For the loss function, id loss and triplet loss are used in TransReID groups. It is worth noting that, for the loss of a batch of samples, each part of loss is the sum of half of the loss value of the first sample and half of the sum of the loss value of the remaining samples. 

\section{Experiment}
\label{sec:experiment}
To form baseline performance, we apply two recent re-identification methods, i.e., AGW \cite{agw} and TransReID \cite{transreid}, to train ASR models on our newly released dataset. In what follows, we firstly introduce our experiment setup and then make an analysis of baseline performance.

\subsection{Setup}
We apply random horizontal flip, padding, random crop, normalization, and random erasing on training data. Mini-batch stochastic gradient descent are used as optimizer. During the training phase, each mini-batch contains 16 roles and each role has four images. The training epoch number is 100. Due to AGW and TransReID being two different models, their learning rates setting are different. The multi-step learning rate scheduler with warm-up is applied for AGW. The learning rate started at $ 1 \times 10^{-4} $ and rise proportionally to $ 1 \times 10^{-2} $ as setting in the first ten epochs of AGW. At the epoch of the $ 40^{th} $  and $ 80^{th} $, the learning rate will drop to one-tenth of the present value. The cosine learning rate scheduler for TransReID with the margin sets to 0.5 and the scale sets to 30. The learning rate started at a third and rise proportionally to $ 8 \times 10^{-3} $ as setting in the first five epochs of TransReID. For the rest of settings, we follow AGW \footnote{The code of original AGW can be found here \url{https://github.com/mangye16/ReID-Survey}.} and TransReID \footnote{The code of original TransReID can be found here \url{https://github.com/damo-cv/TransReID}.}. 

\subsection{Result and Analysis}
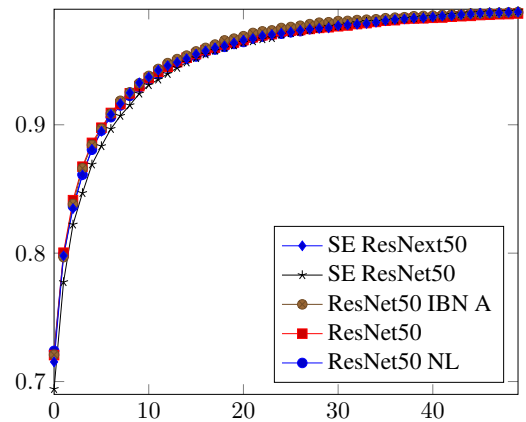
\begin{figure}[t]
    \centering
        \begin{tikzpicture}[scale=0.9]
            \begin{axis}[ymin=0.69, ymax=0.99, xmin=0, xmax=49,
                legend style={cells={anchor=west}, legend pos= south east},
                reverse legend=true,
            ]
                \addplot table [x=rank, y=value, col sep=comma] {CMC-Curve/resnet50_nl.csv};]
                \addlegendentry{ResNet50 NL}
                \addplot table [x=rank, y=value, col sep=comma] {CMC-Curve/resnet50.csv};]
                \addlegendentry{ResNet50}
                \addplot table [x=rank, y=value, col sep=comma] {CMC-Curve/ResNet50_ibn_a.csv};]
                \addlegendentry{ResNet50 IBN A}
                \addplot table [x=rank, y=value, col sep=comma] {CMC-Curve/se_resnet50.csv};]
                \addlegendentry{SE ResNet50}
                \addplot table [x=rank, y=value, col sep=comma] {CMC-Curve/se_resnext50.csv};]
                \addlegendentry{SE ResNext50}
            \end{axis}
        \end{tikzpicture}
    \caption{CMC curve of several experiments on AGW.}
    \label{fig:agw_result}
\end{figure}

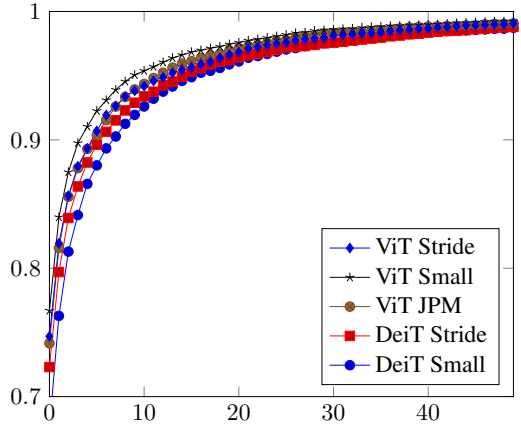
\begin{figure}[t]
    \centering
    \begin{tikzpicture}[scale=0.9]
        \begin{axis}[ymin=0.7, ymax=1, xmin=0, xmax=49,
            legend style={cells={anchor=west}, legend pos= south east},
            reverse legend=true,
        ]
            \addplot table [x=rank, y=value, col sep=comma] {CMC-Curve/deit_small.csv};
            \addlegendentry{DeiT Small}
            \addplot table [x=rank, y=value, col sep=comma] {CMC-Curve/deit_stride.csv};
            \addlegendentry{DeiT Stride}
            \addplot table [x=rank, y=value, col sep=comma] {CMC-Curve/vit_jpm.csv};
            \addlegendentry{ViT JPM}
            \addplot table [x=rank, y=value, col sep=comma] {CMC-Curve/vit_small.csv};
            \addlegendentry{ViT Small}
            \addplot table [x=rank, y=value, col sep=comma] {CMC-Curve/vit_stride.csv};
            \addlegendentry{ViT Stride}
        \end{axis}
    \end{tikzpicture}
    \caption{CMC curve of several experiments on TransReID.}
    \label{fig:transreid_result}
\end{figure}

Following the above-mentioned cross-role protocol, we evaluate the AGW model with different backbones (e.g., ResNet50 \cite{resnet}, ResNet NL \cite{nonlocal}, ViT \cite{vit} and DeiT \cite{deit}) on LSASRD. The results of AGW and TransReID are shown in \Cref{tab:agw} and in \Cref{tab:transreid}, respectively. The CMC cures of AGW and TransReID shown in \Cref{fig:agw_result} and \Cref{fig:transreid_result}, respectively. From \Cref{tab:agw}, we can see that the ResNet50 gets the best mINP and mAP, while ResNet50 NL gets the best Rank1 and Rank5. \Cref{fig:agw_result} more comprehensively that variations of ResNet50 resulted in the tiny performance difference. And with the increasing of the number of layers, the performance does not increase accompanying. In a possible consideration, with the same configuration of hyper-parameters, deeper models do not give full play to their due performance. According to \Cref{tab:transreid}, ViT-stride model get the best mINP, while ViT-small get the best mAP, Rank1 and Rank5. We can reasonably doubt that the ViT model pretrained on a huge amount of concrete images needs to be adjusted to fit our abstract images dataset.

\begin{table}[t]
    \centering
    \caption{Comparisons of performance on AGW with different backbones on LSASRD.}
    \begin{tabular*}{0.98\columnwidth}{@{}@{\extracolsep{\fill}}lcccc@{}}
        \toprule
        Metrics & mINP & mAP & R1 & R5 \\
        \midrule
        ResNet50 \cite{resnet} & \textbf{12.48} & \textbf{40.84} & 72.06 & 88.60 \\
        ResNet50 NL \cite{nin} & 12.40 & 40.80 & \textbf{72.50} & \textbf{88.18} \\
        ResNet101 \cite{resnet} & 12.30 & 40.18 & 70.78 & 87.28 \\
        ResNet152 \cite{resnet} & 12.26 & 40.34 & 71.82 & 88.04 \\
        SE ResNet50 \cite{senet} & 10.76 & 38.10 & 69.44 & 86.90 \\
        SE ResNet101 \cite{senet}& 10.86 & 38.42 & 70.10 & 86.86 \\
        SE ResNet152 \cite{senet}& 10.28 & 36.80 & 67.26 & 86.10 \\
        SE ResNext50 \cite{resnext} & 10.90 & 39.44 & 71.52 & 88.10 \\
        SE ResNext101 \cite{resnext} & 9.32 & 37.56 & 71.54 & 88.52 \\
        ResNet50 IBN A \cite{ibna} & 10.90 & 40.74 & 71.52 & 88.10 \\
        \bottomrule
    \end{tabular*}
    \label{tab:agw}
\end{table}

\begin{table}[t]
    \caption{Comparisons of performance on TransReID with different Transformer settings on LSASRD.}
    \centering
    \begin{tabular*}{0.98\columnwidth}{@{}@{\extracolsep{\fill}}lcccc@{}}
        \toprule
        Metrics & mINP & mAP & R1 & R5\\
        \midrule
        DeiT-Small \cite{deit} & 10.58 & 36.58 & 67.54 & 86.56 \\
        DeiT-Stride \cite{deit} & 11.34 & 39.66 & 72.32 & 88.22 \\
        ViT-Small \cite{vit} & 12.48 & \textbf{42.76} & \textbf{76.68} & \textbf{91.04} \\
        ViT-Base \cite{vit} & 11.34 & 36.70 & 65.70 & 82.98 \\
        ViT-JPM \cite{transreid} & 12.72 & 41.88 & 74.16 & 89.30 \\
        ViT-Stride \cite{vit} & \textbf{13.14} & 42.24 & 74.72 & 89.34 \\
        \bottomrule
    \end{tabular*}
    \label{tab:transreid}
\end{table}

We found no significant difference in experimental results of AGW based model and TransReID based model. It could be the reason that the configuration of models above focuses on images of natural persons. Current models have a good performance in extracting the texture feature of an image. However, they lack a mechanism for learning the semantics information in an image. We need a more specific configuration to explore the capacity limit of models.

\begin{figure}[t]
    \centering
    \begin{minipage}[t]{\columnwidth}
        \centering
        \includegraphics[width=0.24\textwidth]{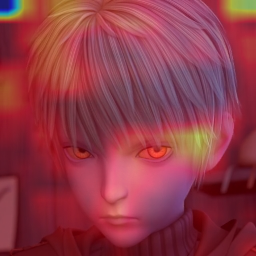}
        \includegraphics[width=0.24\textwidth]{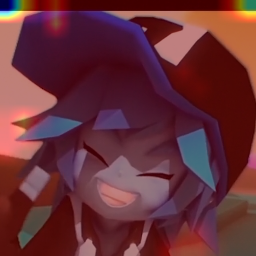}
        \includegraphics[width=0.24\textwidth]{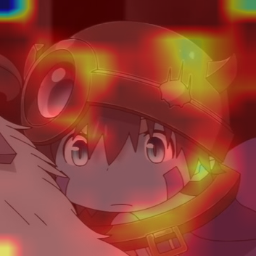}
        \includegraphics[width=0.24\textwidth]{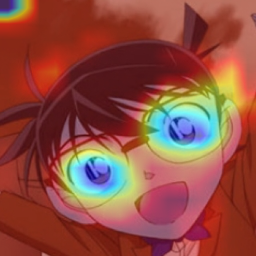}
        \subcaption{AGW}
    \end{minipage}
    \begin{minipage}[t]{\columnwidth}
        \centering
        \includegraphics[width=0.24\textwidth]{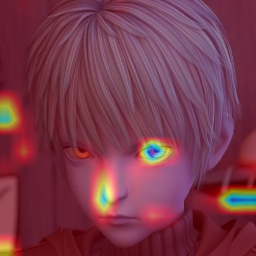}
        \includegraphics[width=0.24\textwidth]{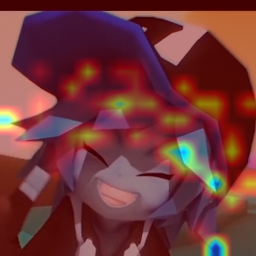}
        \includegraphics[width=0.24\textwidth]{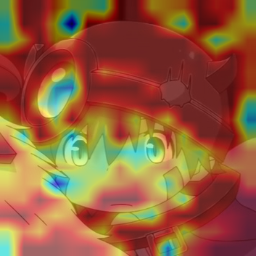}
        \includegraphics[width=0.24\textwidth]{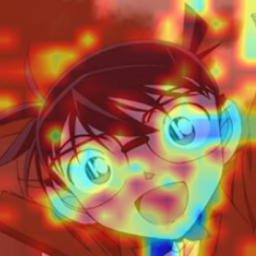}
        \subcaption{TransReID}
    \end{minipage}
    \caption{Heat map \cite{heatmap} of samples. Samples are numbered as 1,2,3 and 4 of both AGW and TransReID from left to right.}
    \label{fig:HMS}
\end{figure}

Based on GradCam++ \cite{gradcampp}, we visualize the gradient and activation of the last Bottleneck block of layer4 module of AGW using the ResNet50 NL backbone and the first LayerNorm layer of the first block in b1 module of TransReID using ViT. From \Cref{fig:HMS}, we can observe that, for the first two 3D anime image samples, few features are noticed by AGW while TransReID notices the edges and shadows. For the last two samples, both models are attractive in the significant features of the roles, Kabuto of the third sample and the big and unique drawing style of eyes of the fourth sample. While TransReID mentioned more parts of the sample than AGW.

\section{Conclusion}

\label{sec:con}
This paper presents a challenging anime style recognition (ASR) benchmark, which explores the semantic understanding capability of the deep learning model. Concurrently, to promote research in the field of anime-style images, we construct a large-scale ASR dataset (LSASRD). With 20937 images of 1829 roles in 190 works, LSASRD contains rich metadata of every label and all images collected handcrafted. We design a cross-role protocol to evaluate ASR performance. Conducting different experiments on the LSASRD with two SOTA re-identification methods, we report the baseline performance on various models. Experiment results show that SOTA re-identification methods cannot achieve satisfactory performance on the ASR benchmark of all, and it shows that current methods are inadequate to extract abstract attributes, i.e., the features of between huge semantic gap. ASR provides a framework for research both the ability of semantic understanding of model and application of anime images.

\section*{Acknowledgements}
This work was supported by National Natural Science Foundation of China under the Grants 61976098 and 61871434; National Key R\&D Program of China under the Grant 2021YFE0205400; Natural Science Foundation for Outstanding Young Scholars of Fujian Province under the Grant 2019J06017; Collaborative Innovation Platform Project of Fuxiaquan National Independent Innovation Demonstration Zone under the Grant 2021FX03; National Training Program on Undergraduate Innovation and Entrepreneurship of China. 

{\small
\bibliographystyle{ieee_fullname}
\bibliography{egbib}
}

\end{document}